\pdfoutput=1

\documentclass[11pt]{article}

\usepackage{ACL2023}

\usepackage{times}
\usepackage{latexsym}

\usepackage[T1]{fontenc}

\usepackage[utf8]{inputenc}

\usepackage{microtype}

\usepackage{inconsolata}
\usepackage{hyperref}
\usepackage{url}
\usepackage[utf8]{inputenc} 
\usepackage[T1]{fontenc}    
\usepackage{hyperref}       
\usepackage{url}            
\usepackage{booktabs}       
\usepackage{amsfonts}       
\usepackage{nicefrac}       
\usepackage{microtype}      
\usepackage{xcolor}         
\usepackage{amsmath,amsfonts,bm}
\usepackage[nameinlink]{cleveref}

\crefformat{section}{\S#2#1#3} 
\crefname{algorithm}{Alg.}{Algs.}
\crefformat{subsection}{\S#2#1#3}
\Crefname{equation}{Eq.}{Eqs.}
\Crefname{figure}{Fig.}{Figs.}

\newcommand{\ie}{{\em i.e.,}\xspace}
\newcommand{\eg}{{\em e.g.,}\xspace}

\newcommand{\Ni}{({\em i})~}
\newcommand{\Nii}{({\em ii})~}
\newcommand{\Niii}{({\em iii})~}

\makeatletter   
\newcommand{\sveryshortarrow}[1][3pt]{\mathrel{%
    \vcenter{\hbox{\rule[-.5\fontdimen8\scriptfont3]
               {\scriptratio\dimexpr#1\relax}{\fontdimen8\scriptfont3}}}%
   \mkern-4mu\hbox{\let\f@size\sf@size\usefont{U}{lasy}{m}{n}\symbol{41}}}}
\makeatother

\def\eqref#1{equation~\ref{#1}}

\def\1{\bm{1}}

\def\rvc{{\mathbf{c}}}

\def\rvx{{\mathbf{x}}}

\def\vzero{{\bm{0}}}

\def\m1{{\bm{1}}}

\DeclareMathAlphabet{\mathsfit}{\encodingdefault}{\sfdefault}{m}{sl}
\SetMathAlphabet{\mathsfit}{bold}{\encodingdefault}{\sfdefault}{bx}{n}

\def\gB{{\mathcal{B}}}
\def\gC{{\mathcal{C}}}
\def\gD{{\mathcal{D}}}

\def\gL{{\mathcal{L}}}

\def\gR{{\mathcal{R}}}
\def\gS{{\mathcal{S}}}

\usepackage[pdftex]{graphicx}
\usepackage{wrapfig}

\usepackage{enumitem}
\usepackage{bbm}
\usepackage{amsthm}
\usepackage[colorinlistoftodos,prependcaption,textsize=tiny]{todonotes}
\usepackage{multirow}
\usepackage{soul}
\usepackage{color}
\usepackage{verbatim}
\definecolor{greena}{RGB}{0,255,0} 
\definecolor{greenb}{RGB}{128,255,128} 
\definecolor{greenc}{RGB}{230,255,230} 
\DeclareRobustCommand{\hlgreena}[1]{{\sethlcolor{greena}\hl{#1}}}
\DeclareRobustCommand{\hlgreenb}[1]{{\sethlcolor{greenb}\hl{#1}}}
\DeclareRobustCommand{\hlgreenc}[1]{{\sethlcolor{greenc}\hl{#1}}}

\usepackage{inconsolata}
\usepackage{xspace}
\makeatletter
\DeclareRobustCommand\onedot{\futurelet\@let@token\@onedot}
\def\@onedot{\ifx\@let@token.\else.\null\fi\xspace}

\def\eg{\emph{e.g}\onedot} 
\def\ie{\emph{i.e}\onedot} 
\usepackage{adjustbox}

\title{Explaining Language Models' Predictions with High-Impact Concepts}

\author{%
  Ruochen Zhao$^{1}$ ~~
   Shafiq Joty$^{1,2}$ ~~
   Yongjie Wang$^{1}$ ~~ Tan Wang$^{1}$ \\
  $^1$Nanyang Technological University, Singapore\\
  $^2$Salesforce AI\\
  \{ruochen002, yongjie002, tan317\}@e.ntu.edu.sg \\
  srjoty@ntu.edu.sg \\
}

\begin{document}
\maketitle
\begin{abstract}
The emergence of large-scale pretrained language models has posed unprecedented challenges in deriving explanations of why the model has made some predictions. Stemmed from the compositional nature of languages, spurious correlations have further undermined the trustworthiness of NLP systems, leading to unreliable model explanations that are merely correlated with the output predictions. To encourage fairness and transparency, there exists an urgent demand for reliable explanations that allow users to consistently understand the model's behavior. In this work, we propose a complete framework for extending concept-based interpretability methods to NLP. Specifically, we propose a post-hoc interpretability method for extracting predictive high-level features (concepts) from the pretrained model's hidden layer activations. We optimize for features whose existence causes the output predictions to change substantially, \ie generates a high impact. Moreover, we devise several evaluation metrics that can be universally applied. Extensive experiments on real and synthetic tasks demonstrate that our method achieves superior results on {predictive impact}, usability, and faithfulness compared to the baselines. Our codebase is available at \url{https://github.com/RuochenZhao/HIConcept}.
\end{abstract}


\section{Introduction}

Over the past few years, progress achieved by large language models (LLMs) has led them to be widely applied in sensitive applications such as personalized recommendation bots and recruitment. However, many users are still reluctant to adopt large-scale NLP models due to their opaque decision processes \citep{mathews2019explainable}. To increase transparency and user trust, it is crucial to utilize Explainable AI (XAI) to derive effective model interpretations. Following interpretability definitions proposed by \citet{miller2019explanation} and \citet{kim2016examples}, we hope to allow humans to understand the cause of a model prediction, thus increasing the degree to which a human can consistently predict the model’s results.

\begin{figure}[t!]
    \centering
    \includegraphics[width=0.5\textwidth]{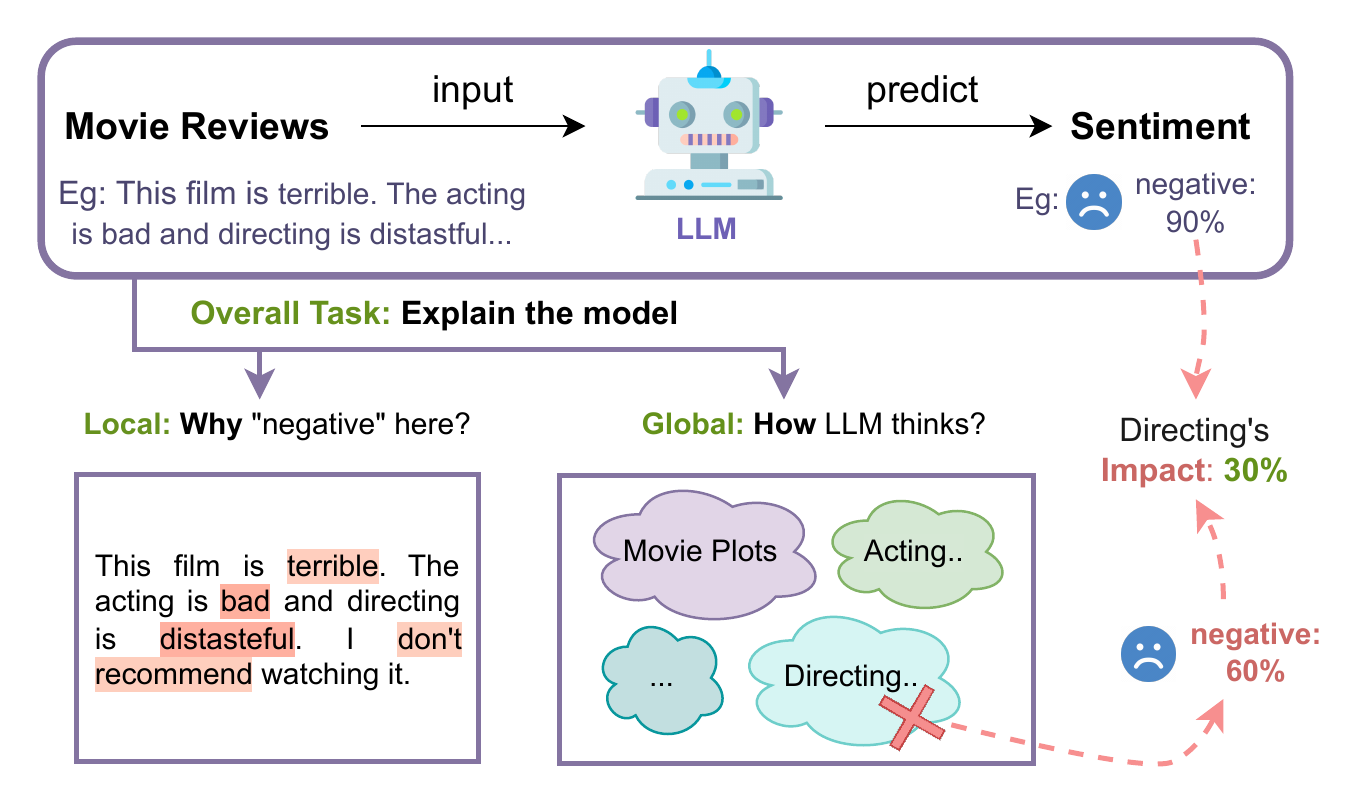}
  \caption{\small Illustration of local and global explanation. On the right, impact of the concept ``Directing'' is the change in prediction after its removal.}
  \label{fig:intro_fig}
\end{figure}
Current XAI methods fall into two categories based on model scope: \Ni local or instance-based, and \Nii global or model-based \citep{adadi2018peeking}.
Illustrated in \Cref{fig:intro_fig}, to explain a sentiment classification model, local explanations focus on answering ``why the model predicts negative for this particular instance''. Local methods such as counterfactuals have been widely used to find the minimal change to an instance that results in an alternative prediction \cite{s2017counterfactual,wu2021polyjuice}. Global explanations, on the other hand, attempt to explain the decision process at the model level, such as using feature clusters like `movie plot' and `acting'. Among them, concept-based models, such as \citet{koh2020concept} and \citet{kim2018interpretability}, extract high-level concepts to characterize the global behavior of deep neural models. When understanding texts, humans often employ \emph{concept-based reasoning}, such as grouping words into topics \citep{NIPS1998_Tenenbaum}. Therefore, concept-based interpretability provides a natural way of deriving explanations for NLP models.

However, one shortcoming of the existing concept-based methods is the lack of consideration of the explanations' ``impact'' on output predictions. We define the ``impact'' of a feature to be the change in output predictions without it. For example, the impact of the \emph{directing} concept in \Cref{fig:intro_fig} is the change in prediction after removing it. Current concept-based methods emphasize learning features that can recover a model's predictions accurately. Thus, these features highly correlate with the predictions. However, as correlation does not always imply causation, we find that such explanations often have low impact, \ie excluding it will not cause the predictions to be different. This undermines the validity of such explanations as users cannot utilize them to consistently predict the model's behavior when a feature changes.  This is especially concerning for NLP. Stemming from the compositional nature of languages, LLMs contain a large number of \emph{spurious correlations} -- features that are useful for training but not causal, which have become a serious threat \citep{causalNLP,mccoy-etal-2019-right, eisenstein2022uninformative}. Therefore, simply applying correlation-based concept mining methods is problematic, as it may often generate correlational explanations.



In this work, we propose \emph{HI-Concept}, a complete framework for explaining LLMs based on high-level concepts that directly impact the model's predictions by loss construction. As a post-hoc approach, our method trains a lightweight concept model that discovers latent features from hidden activations as global concepts, which can also be mapped back to generate local explanations. Specifically, to optimize for a high-impact set of concepts, we design a \emph{causal loss} to be optimized during training the concept model. To evaluate the faithfulness, usability, and impact of the discovered concepts, we devise several evaluation metrics and conduct human studies. In our experiments {with 2 classification datasets and 3 pretrained models}, we observe that HI-Concept consistently finds high-impact concepts that are more faithful and usable. Our contributions can be summarized as follows:

\begin{itemize}[leftmargin=*,topsep=2pt,itemsep=2pt,parsep=0pt]
\item We first propose HI-Concept, a method to derive both global concepts and their corresponding local explanations that result in high output changes. Both forms of generated explanations can complement each other while conforming to the `mindset' of the model.
\item We propose evaluation metrics that {stem from} theoretical definitions of treatment effects in literature \citep{reason:Pearl09a} and design a human study to prove the usability of the derived explanations.
\item We show that HI-Concept is impactful, usable, and faithful by constructing reliable and extensive experiments.
\end{itemize}



\section{Related Work}

\subsection{Explainability methods}


\paragraph{Concept-based (General-domain):}  

Concept-based methods \cite{kim2018interpretability} have been a popular group of interpretability methods as it derives user-friendly, high-level concepts as explanations. 
Among them, the most recent method is ConceptShap \cite{yeh2020completeness}, which discovers concepts in the intermediate layer with a bottleneck-shaped extraction model and proposes an adapted Shapley value metric to evaluate completeness scores. 
However, as we will show in our experiments (\Cref{sec:exp}), because the existing concept-based methods do not differentiate between correlational and causal information, their performances on NLP tasks is problematic: the discovered concepts often have little impact on final model predictions. Especially on complex transformer models with stronger confounding effects brought by pretraining, their performances may further decrease.

\vspace{-0.5em}
\paragraph{NLP:} 

Many NLP explainability methods only discover correlational features, such as induction-based methods \cite{ling2017program}, explainability-aware architectures \cite{rajani2019explain}, and feature importance scores \cite{croce2019auditing}.


\subsection{Causal explainability methods}

\paragraph{General-domain} \citet{harradon2018causal} attempt to intervene in an unsupervised way on the hidden space by constructing several even-spaced Variational Autoencoders (VAEs) throughout a CNN, but they only train with a reconstruction loss instead of explicitly optimizing for {impact}. 
Probing methods \citep{conneau2018you,belinkov-etal-2020-linguistic} train an external model - a \emph{probe} - to predict some properties of interest from the latent representations. To further investigate causal effects of the features learned from probing, \citet{elazar2021amnesic} assess the influence of a causal intervention by removing a feature. However, subsequent work \cite{barrett2019adversarial} shows that such methods generalize poorly to unseen samples. Moreover, as \citet{belinkov2022probing} points out, the disconnect between the probing model and the original model may result in the properties not being utilized in the original model's prediction task. Causal Mediation Analysis (CMA) \cite{pearl2022direct} measures the change in an output following a counterfactual intervention in an intermediate variable, or mediator.

\vspace{-0.5em}
\paragraph{NLP} According to \citet{causalNLP}, causality shows a promising path forward for NLP research, which can offer insights into the model's inner workings. Most current methods attempt to causally explain an NLP model by generating \emph{counterfactual} inputs \cite{alvarez2017causal,veitch2021counterfactual,wu2021polyjuice}. 
\citet{vig2020investigating} apply CMA to examine gender bias by changing pronouns in the input. We argue that both probing and CMA could be limited to a few aspects as they rely on human-constructed features (\eg , linguistic, gender features), requiring expertise on the datasets and tasks. Thus, it might be beneficial to develop unsupervised explanation features.

\section{Methodology}

\subsection{Task formulation}

We follow the definition of interpretability by \citet{kim2016examples}: ``the degree to which a human can consistently predict the model’s result.'' Given a pretrained model $f$, we aim to explain why it makes a certain predictions $f(\rvx)$ for an input $\rvx$. To ensure validity, {the explanation should} significantly contribute to the overall prediction process, thus enabling humans to ``consistently predict'' the results. As {$f$ is fixed, its} prediction process is deemed deterministic and reproducible, allowing us to conduct counterfactual experiments.

We generally follow the setup in traditional concept extraction models \citep{yeh2020completeness}, where explanations are high-level features (concepts) discovered on the hidden activation space with a concept extraction model. The pretrained model {$f$} can be viewed as a composite of two functions, divided at an intermediate layer {$f = \psi \circ \phi$}. 
$\phi(\cdot)$ {encodes} the input text $\rvx$ to a hidden representation $\phi(\rvx)$, and 
$\psi(\cdot)$ maps $\phi(\rvx)$ to classification probabilities $\psi(\phi(\rvx))$. Without loss of generality, we assume that, for an input  $\rvx$, which consists of $T$ tokens $[x_1, \dots, x_T]$, $\phi(\rvx)$ can be represented as a concatenation of $[\phi(x_1), \ldots, \phi(x_T)]$, where each $\phi(x_t) \in \mathbb{R}^d$ denotes a representation of an input token $x_t$. Depending on the  model architecture, $\phi(x_t)$ can be encoded from a local receptive field as in convolutional nets or a global one as in Transformers \citep{vaswani2017attention}. 
As the hidden space $\phi(\rvx)$ reflects the input distribution $\rvx$, $n$ concepts $\gC = \{\rvc_1, \dots, \rvc_n\}$ {could be extracted on $\phi(\rvx)$} that {correspond to} different {input} features.

\subsection{Correlated VS. predictive explanations}

\begin{figure}[t!]
    \centering
    \includegraphics[scale=0.35]{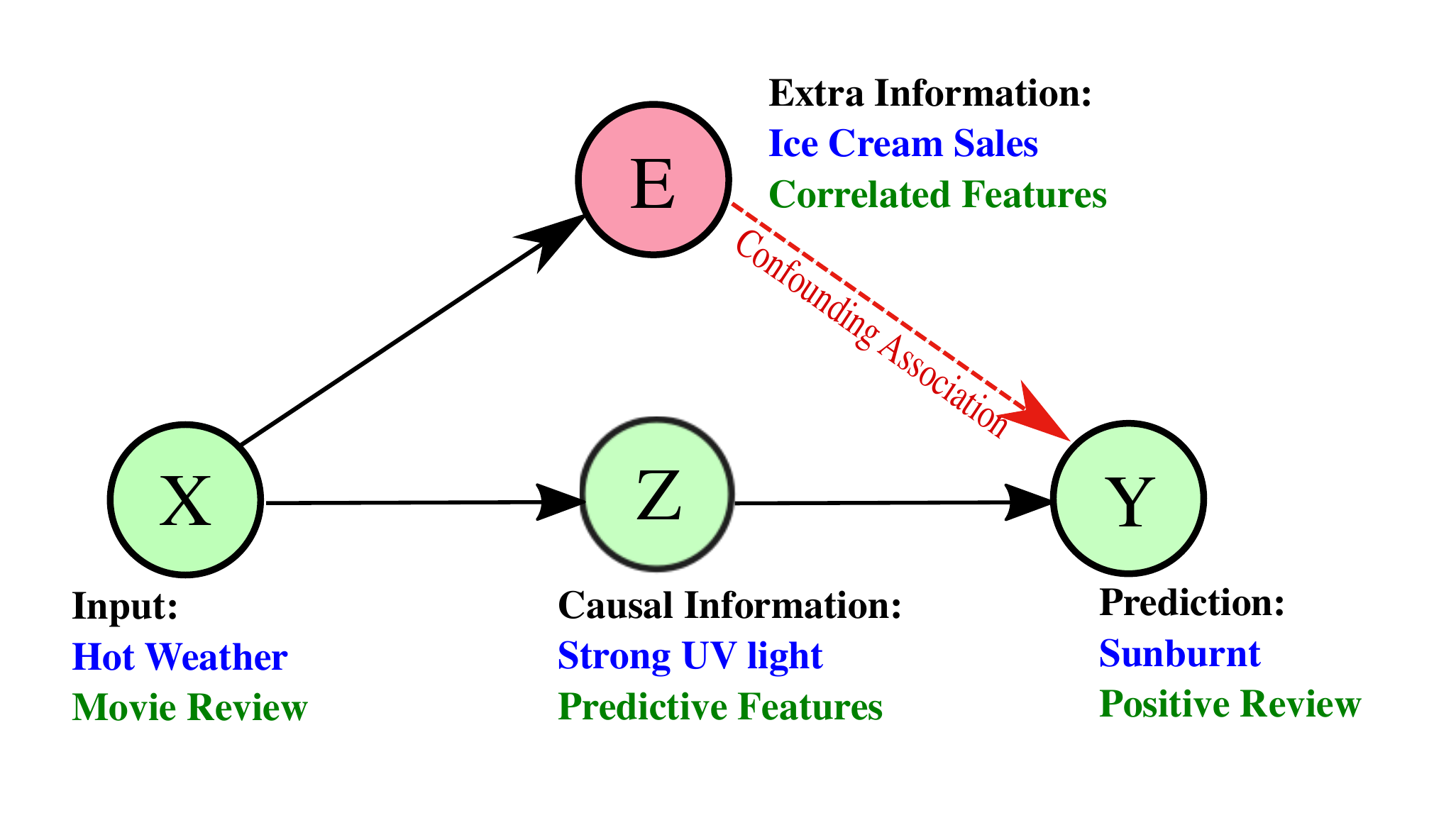}
   \vspace{-1em}
  \caption{Causal graph illustration. }
  \label{fig:causal_graph}
\end{figure}

The general approach to concept-based explanation has been to use a concept mining model to discover high-level features from $\phi(\rvx)$.
However, it only finds 
\emph{correlational} features, as it does not force explanations to be used in final predictions. The failure cases can be explained by drawing inspiration from causality analysis (\Cref{fig:causal_graph}). A real-life analogy is that, while the hot weather ($X$) creates high demand for ice cream ($E$), it also produces intense UV light exposure ($Z$), causing more sunburns ($Y$). However, it is obvious that high ice cream sales ($E$) do not cause sunburns ($Y$).

In pretrained LLMs, the hidden activation space consists of both $E$ and $Z$. Although only $Z$ truly affects prediction $Y$, $E$ and $Z$ may be correlated due to the confounding effects brought by $X$. However, a traditional concept mining model does not differentiate between them and considers both as valid, which is problematic. In our experiments  (\Cref{sec:exp}), we also observe that the discovered concepts based on current methods often have little to no impact on output predictions, \ie generate tiny changes in predictions when a concept is turned off, especially when interpreting LLMs like BERT \citep{devlin2018bert} and T5 \citep{raffel2020exploring}. 
{To tackle this challenge}, we {enforce} explanations to {be predictive by considering their ``impact''.}\footnote{We assume that, as the concept vectors coexist in the hidden embedding space of $E \cup Z$, there is no causal relationship among the concepts $\{\rvc_1, \dots, \rvc_n\}$ themselves.}

\subsection{Defining impact}
To formally define the \emph{impact} of a feature, we draw inspiration from causal analysis, namely Individual Treatment Effect (ITE) and Average Treatment Effect (ATE), which measure the effect of interventions in randomized experiments. Given a binary treatment variable $T$ that indicates whether an intervention is performed, ATE and ITE are defined as the change in expected outcome with the do-operation:
\begin{equation}
\small 
\begin{split}
\text{ITE}(x) :=&  ~~\mathbb{E}[y|\textbf{X}=x, \text{do}(T=1)] \\
& ~~~~  - \mathbb{E}[y|\textbf{X}=x, \text{do}(T=0)]; \\
\text{ATE} :=& ~~\mathbb{E}[\text{ITE}(x)]
\end{split}
    \label{eq:ATE_ind}
\end{equation}

In our case, a concept $\rvc_i$ is discovered as a direction in the latent space, corresponding to a feature in the input distribution. If $\rvc_i$ is used by the model for prediction, the removal of $\rvc_i$ should cause a change in prediction. 
Thus, we define \emph{impact} I of a concept $\rvc_i$ on an instance $(\rvx, y)$ as:
\begin{equation}
\small 
\begin{split}
\text{I}(\rvc_i, \rvx)  =  \mathbb{E}[y|\textbf{X}=\rvx, \rvc_i = \vzero]
 - \mathbb{E}[y |\textbf{X}=\rvx, \rvc_i = \rvc_i]
\label{eq:impact}
\end{split}
\end{equation}
Then, \emph{individual impact} ($\text{I}(\rvc_i))$ of a concept $\rvc_i$ and \emph{average impact} ($\text{I}(\gC)$) for a concept set $\gC$ could be written as:
\begin{equation}
\small 
\begin{split}
\text{I}(\rvc_i) := \mathbb{E}[\text{I}(\rvc_i, \rvx)]; ~~\hspace{1ex} \text{I}(\gC) =& \mathop{\mathbb{E}}_{\rvc_i \in  \gC}[\text{I}(\rvc_i)]
\end{split}
\label{eq:II_and_AI}
\end{equation}

\subsection{Concept vector extraction}
 
We approximate the distribution over the concepts by encoding the hidden activations $\phi(x_t) \in \mathbb{R}^d$ into concept probabilities $p_{\gC}(x_t) = [p_c^1(x_t), \dots, p_c^n(x_t)]$ with a bottleneck-shaped network while optimizing for \emph{impact}. For a visualization of the overall model architecture, we refer readers to \Cref{app:overall_graph}.

\paragraph{Utilizing the original concept model}

\noindent After initializing the concepts $\gC = \{\rvc_1, \dots, \rvc_n\}$ uniformly, 
{the probability for concept $\rvc_i$} is calculated as $p_c^i(x_t) = \text{TH}((\phi(x_t)^{\top} \rvc_i), \beta)$, where $\text{TH}$ is a threshold function that forces all inputs smaller than $\beta$ to be $0$. 
To get the concept distribution for the entire sequence $\rvx$, we concatenate the token-level distributions: $p_{\gC}(\rvx) = [p_{\gC}(x_1), \ldots, p_{\gC}(x_T)] \in \mathbb{R}^{T\times n}$. For transformer-based models such as BERT and T5, $\phi(\rvx)$ consists of only the sequence-level representation (e.g., \texttt{[cls]}) that is used for prediction.
{After encoding $\phi(\rvx)$ into $p_{\gC}(\rvx)$, the bottleneck-shaped network} reconstructs $\phi(\rvx)$ with a 2-layer perceptron $g_{\theta}$ such that $g_{\theta}(p_{\gC} (\rvx)) \approx \phi(\rvx)$. 

Then, two loss terms are proposed to train the concept model in an end-to-end way:
     
\noindent \textit{$\bullet$ Reconstruction loss:} To faithfully recover the original DNN model's predictions, we optimize a surrogate loss with cross-entropy (CE) defined as:
\begin{equation}
\small 
\begin{split}
    \gL_{\text{rec}}{(\theta, \gC)} & = \text{CE}\Big( \psi \big(\phi({\rvx}) \big), \psi \big( g_{\theta}(p_{\gC}(\rvx)) \big) \Big) \\ 
    & = - \sum_{{b} \in \gB} \psi \big( \phi(\rvx) \big)_b \log \big( \psi(g_{\theta}(p_{\gC}(\rvx)))_b \big)
\end{split}
\label{eq: pred_const_loss}
\end{equation}   

\noindent where $\gB$ is the set of class labels and $\psi ( . )_b$ denotes the prediction score corresponding to label $b$.  

\noindent \textit{$\bullet$ Regularization loss:} To ensure user-friendliness, we {encourage} each concept vector {to }correspond to actual examples and concepts {to be} distinct from each other. 
Formally,  
\begin{equation}
\small 
\begin{split}
    \gL_{\text{reg}}(\gC) = & - \lambda_1 \frac{\sum_{i=1}^n \sum_{x_t \in \gR_{i}} \rvc_i^{\top} \phi(x_t)}{nN} \\
    & + \lambda_2 \frac{\sum_{i_1 \neq i_2} \rvc_{i_1}^{\top} \rvc_{i_2}}{n(n-1)}
\end{split}
\label{eq:reg_loss}
\end{equation}


\paragraph{Optimizing for impact}

{We introduce two losses to the original framework: \Ni Auto-Encoding loss, which guarantees that the discovered concepts can serve as latent features representing the original distribution of $\phi({\rvx})$; \Nii Causality Loss, which mimics an estimation of ``impact'' proposed in \Cref{eq:II_and_AI}. By directly optimizing these two new objectives, HI-Concept ensures that the discovered concepts generate a high impact on the prediction process.}

\noindent \textit{$\bullet$ Auto-encoding loss:} 
we define the following mean-squared error (MSE) loss for auto-encoding:
\begin{equation}
\small 
\begin{split}
    \gL_{\text{enc}}{(\theta, \gC)} & = \text{MSE}\Big(  \phi({\rvx}), g_{\theta}(p_{\gC}(\rvx)) \Big) \\
    & = \frac{1}{d}||\phi({\rvx}) - g_{\theta}(p_{\gC}(\rvx)||_2^2
\end{split}
\label{eq: pred_ae_loss}
\end{equation}   



\noindent \textit{$\bullet$ Causality loss:} This loss is designed to disentangle concept directions that have a greater impact. Following \Cref{eq:II_and_AI}, our intuition is that a merely correlational concept should have {an impact} close to $0$. Therefore, we optimize the following loss.
\begin{equation}
\small 
\begin{split}
    \gL_{\text{cau}}(\theta, \gC) =  & - \hspace{-0.5em} \mathop{\sum}_{\rvc_i \in \gS} \mathop{\sum}_{\rvx_j \in \gD} \Big| \psi \Big( g_\theta(p_{\gC}(\rvx_j) | \rvc_i = \vzero) \Big) \\
    & - \psi \Big( g_\theta(p_{\gC}(\rvx_j) | \rvc_i = \rvc_i) \Big) \Big| 
    \approx - |I_\text{avg}(\gC)|
\end{split}
\label{eq: causal_loss}
\end{equation}
Here, $\gS \subseteq \gC$ denotes a set of concepts to remove.
\footnote{During experiments, we have found that random selection yields the best performance.} As we perturb on all inputs $\rvx_j \in \gD$, the training dataset $\gD$ serves both as the treatment group and the nontreatment group, ensuring that no divergence is present. 
As the term $| \psi (\cdot|\rvc_i = \vzero) - \psi (\cdot|\rvc_i = \rvc_i)|$  in Eq. \ref{eq: causal_loss}  approximates the {impact $I(\rvc_i, \rvx_j)$} in Eq. \ref{eq:impact},
minimizing the designed causality loss is a close approximation to maximizing the {average impact in Eq. \ref{eq:II_and_AI}}. Intuitively, this loss encourages the concepts to incorporate directions that result in more significant changes in the output predictions.

\paragraph{Total loss for HI-Concept}
Finally, the overall loss function that we minimize becomes:
\begin{equation}
\small 
\begin{split}
\gL(\theta, \gC) = & \gL_{\text{rec}}{(\theta, \gC)} + \gL_{\text{reg}}(\gC) \\
& + \lambda_e \gL_{\text{enc}}(\theta, \gC) + \lambda_c \gL_{\text{cau}}(\theta, \gC) 
\end{split}
\label{eq: overall_loss}
\end{equation}

\noindent where $\lambda_e$, $\lambda_c$ are the weights for the auto-encoding loss and the causal loss respectively. In practice, the hyperparameters require minimal tuning. Specifically, we recommend fixing $\lambda_1=0.1$ and $\lambda_2=0.5$ for regularizer loss in \Cref{eq:reg_loss}, and $\lambda_e = 1$ for reconstruction loss. 
The only hyperparameter to tune is $\lambda_c$, whose optimal level can be found within a few steps. Futher details on implementation and the training process could be found in \Cref{app:training}.


\paragraph{Mapping concepts back to word tokens} \label{subsec:map}

To derive corresponding local explanations, we employ existing techniques to map activation-space features back to the discrete input tokens when the receptive field of a concept is larger than token-level. 
For explaining BERT, we employ the transformer visualization method proposed in \citet{chefer2021transformer} 
to map back from the [CLS] activation concepts to input  tokens. Specifically, it visualizes classification with a combination of layer-wise propagation (LRP), gradient backpropagation, and layer aggregation with rollout.
As a result, for each sample $\rvx$ and concept $\rvc_i$, we go from having only one concept similarity score $p_c^i(\rvx)$ to having a list of normalized token importance scores $s_1(\rvc_i), \dots, s_T(\rvc_i)$. For the intermediate layers of BERT, we simply use the corresponding token representation ({\em i.e.,}\ $\phi(x_t)$ and its corresponding $p_c^i(x_t)$). For CNNs, we employ the GradCam approach \citep{selvaraju2017grad}, which rolls out the gradients to produce scores for each token.



\subsection{Evaluating impact}
\label{subsec:metric}

To evaluate \emph{impact} for a feature, we devise three metrics based on its definition in \Cref{eq:II_and_AI}, which can be globally applied to other feature-based interpretability methods.

\noindent \textit{$\bullet$ Individual and average impacts:} We approximate \Cref{eq:II_and_AI} on the test set $\gD_{\text{test}}$:
\begin{equation*}
\small 
\begin{split}
\text{I}(\rvc_i) & := \mathop{\mathbb{\sum}}_{\rvx_j \in \gD_{\text{test}}} |\psi \big( g_{\theta}(p_\gC(\rvx_j)) \big) - \psi \big( g_{\theta}(p_{\gC\setminus\{i\}}(\rvx_j)) \big)| ; \\
\text{I}(\gC)& = \frac{1}{	\lvert \gC \rvert} \mathop{\sum}_{\rvc_i \in \gC} \text{I}(\rvc_i)
\end{split}
\end{equation*}
\noindent \textit{$\bullet$ Recovering accuracy change:} 
Intuitively, if a concept $\rvc_i$ is a crucial factor used by the model to make predictions, omitting it will disrupt the ability to faithfully recover predictions.
Denoting 
$\text{Acc}(\gC)=\text{Accuracy}(\psi \big( g_{\theta}(p_\gC(\rvx)) \big), \phi(\psi(\rvx))$: 
\begin{equation*}
\small 
\begin{split}
    \Delta \text{Acc}_i & = |\text{Acc}(\gC) - \text{Acc}(\gC\setminus\{\rvc_i\})|; \\
    \Delta \text{Acc} & = \frac{1}{\lvert \gC \rvert} \mathop{\sum}_{\rvc_i \in \gC}\Delta \text{Acc}_i
\end{split}
\end{equation*}

\noindent \textit{$\bullet$ Impact score:} To measure the impact of an individual text token $x_t$, we take the top-3 most similar concepts $\gC_{\text{top}}$ to the input $\rvx$ using the normalized similarity score $p_{\gC}(\rvx)$. For each concept $\rvc_i \in \gC_{\text{top}}$, the transformer visualization method (\Cref{subsec:map}) produces normalized token importance scores $\{s_1(\rvc_i), \dots, s_\text{T}(\rvc_i)\}$. The \textbf{impact} score for a token $x_t$ is  defined as: 
\begin{equation*}
\small
\begin{split}
\text{I}(x_t) = \mathop{\sum}_{c_i \in \gC_{\text{top}}} p_c^i(\rvx) s_t(\rvc_i)
\end{split}
\end{equation*}


\section{Experiment Settings} 
\subsection{Datasets and classification models}

We test the effectiveness of our method with two standard text classification datasets: IMDB \citep{maas-EtAl:2011:ACL-HLT2011} and AG-news \citep{Zhang2015CharacterlevelCN}. The IMDB dataset consists of movie reviews labeled with positive or negative sentiments. The AG-news dataset consists of news articles categorized with 4 topics. \Cref{table:datasets} in \Cref{app:hyperparameters} gives a dataset summary. We explain three classification models: \Ni a {$6$-layer} transformer {encoder} trained from scratch, \Nii a pre-trained BERT with finetuning, \Niii a pre-trained T5 \citep{raffel2020exploring} with finetuning.

\begin{table*}[t]
  \centering
\scalebox{0.7}{\begin{tabular}{llccccccc}
    \toprule
    Dataset & Model & Cls.Acc & Metric  & $\beta$-TCVAE  & K-means & PCA & ConceptSHAP & HI-Concept \\
    \toprule
    \multirow{9}{*}{IMDB} & \multirow{3}{*}{Transformer} & \multirow{3}{*}{81.74\%} & $\text{I}(\gC)$ & 0.037 & 0.047 & 0.001 & 0.031 & \textbf{0.150} \\
    & & & $\Delta$Acc & 1.24\% & 2.59\% & 0.01\% & 1.30\% & \textbf{11.06\%} \\
    & & & RAcc & 52.08\% & 83.64\% & 85.18\% & 84.36\% & \textbf{88.78\%}\\
    \cmidrule{2-9}
    & \multirow{3}{*}{BERT} & \multirow{3}{*}{89.14\%} & $\text{I}(\gC)$ & 0.057 & 0.038 & 0.002 & 0.050 & \textbf{0.104} \\
    & & & $\Delta$Acc & 4.10\% & 1.56\% & 0.02\% & 0.06\% & \textbf{9.47\%} \\
    & & & RAcc & 93.86\% & \textbf{98.69\%} & 96.68\% & 95.84\% & 94.53\%\\
    \cmidrule{2-9}
    & \multirow{3}{*}{T5} & \multirow{3}{*}{72.98\%} & $\text{I}(\gC)$ & 0.000 & 0.025 & 0.000 & 0.000 & \textbf{0.094} \\
    & & & $\Delta$Acc & 0.00\% & 1.06\% & 0.02\% & 20.21\% & \textbf{38.34\%} \\
    & & & RAcc & 0.00\% & 75.85\% & 98.86\% & 60.20\% & \textbf{99.50\%}\\
    \midrule
    \multirow{9}{*}{AG} & \multirow{3}{*}{Transformer} & \multirow{3}{*}{88.33\%} & $\text{I}(\gC)$ & \textbf{0.049} & 0.044 & 0.000 & 0.000 & 0.045 \\
    & & & $\Delta$Acc & 6.62\% & 0.07\% & 0.03\% & 0.00\% & \textbf{7.12\%} \\
    & & & RAcc & 98.90\% & 98.16\% & \textbf{99.99\%} & 73.01\% & 99.50\%\\
    \cmidrule{2-9}
    & \multirow{3}{*}{BERT} & \multirow{3}{*}{93.75\%} & $\text{I}(\gC)$ & 0.044 & 0.028 & 0.001 & 0.025 & \textbf{0.058} \\
    & & & $\Delta$Acc & 5.32\% & 7.15\% & 0.01\% & 4.44\% & \textbf{10.54\%} \\
    & & & RAcc & 92.30\% & 86.83\% & 99.79\% & 93.46\% & \textbf{99.90\%}\\
    \cmidrule{2-9}
    & \multirow{3}{*}{T5} & \multirow{3}{*}{94.30\%} & $\text{I}(\gC)$ & 0.000 & 0.011 & 0.000 & 0.000 & \textbf{0.054} \\
    & & & $\Delta$Acc & 0.00\% & 1.49\% & 0.01\% & 0.00\% & \textbf{52.20\%} \\
    & & & RAcc & 0.00\% & 24.87\% & 97.38\% & 0.00\% & \textbf{99.46\%}\\
    \bottomrule
  \end{tabular}}
  \vspace{-0.5em}
    \centering
  \caption{Faithfulness (RAcc$\uparrow$) and causality ($\text{I}(\gC)$$\uparrow$, $\Delta \text{Acc}\uparrow$) evaluation of different text classification methods.}
  \label{table:text-results}
\end{table*}

\subsection{Evaluation measures}

We evaluate the explanation methods based on three important aspects as described below.

\noindent\textbf{$\bullet$ Causality:} Causality is an important consideration to evaluate explainability methods, especially where spurious correlations are strong. \citet{doshi2017towards} state: ``Causality implies that the predicted change in output due to a perturbation will occur in the real system''. We evaluate $\text{I}(\gC)$ and $\Delta \text{Acc}$ proposed in \Cref{subsec:metric}, as a higher $\text{I}(\gC)$ and $\Delta \text{Acc}$ represent a higher change in model predictions, thus a more impactful set of concepts. 

\noindent\textbf{$\bullet$ Usability:} Proposed in \citep{doshi2017towards}, an important desideratum of explainability is to make sure that it provides usable information that assists users to accomplish a task. With the concepts generating a high impact on predictions
, we expect that it can enable end-users to better understand the model's reasoning process, thus being useful for debugging and fairness. We include visualizations and human studies to test it qualitatively. 

\noindent\textbf{$\bullet$ Faithfulness:} Faithfulness evaluates whether our surrogate model can accurately mimic the original model's prediction process. In other words, we want to ensure that the captured concept probabilities $p_{\gC}(\rvx)$ can recover the original model's predictions $\psi \big(\phi({\rvx}) \big)$. We report the \emph{recovering accuracy} for the set of concepts $\gC$: 
\begin{equation*}
\small
\begin{split}
\text{RAcc} = \frac{1}{ |\gD_{\text{test}}|} \mathop{\sum}_{\rvx_j \in \gD_{\text{test}}} \mathbbm{1} \Big(\psi \big(\phi({\rvx_j}) \big) = \psi \big( g_{\theta}(p_{\gC}(\rvx_j)) \big) \Big)
\end{split}
\end{equation*}

\subsection{Baselines and hyperparameters}

As fair comparisons to our method, we can only consider \emph{unsupervised} feature discovery algorithms. Thus, we use conceptSHAP \citep{yeh2020completeness} as a baseline. To compare to VAE methods, we include the disentanglement VAE ($\beta$-TCVAE) by \citet{chen2018isolating}. Moreover, we include comparisons to popular non-parametric clustering techniques, including PCA and k-means.

The full list of hyperparameters used for training the HI-Concept model can be found in \Cref{app:hyperparameters}. Briefly, we use the causal coefficient $\lambda_c \in [1, 3]$, depending on the level of confounding within the dataset. During training, perturbation is performed on the most similar concept to the input. All experiments are conducted on the penultimate layer with 10 concepts. The hyperparameters are chosen as an optimal default through grid search. To make the comparison fair, PCA, K-means, and $\beta$-TCVAE also use 10 dimensions to encode.

\section{Results and Analysis}  \label{sec:exp}

To first provide a sanity check for our method, we conduct a toy experiment with a synthetic graphic dataset where the level of confounding can be controlled with ground truth concepts. \Cref{app:toy} gives details of the experiment. The results show that our method discovers concepts that align with human understanding and consistently outperforms the baseline by deriving more impactful features. As confounding levels in the dataset increase, the performance gap also widens.

\begin{table*}[t]
\centering
  \scalebox{0.7}{  \begin{tabular}{ccl}
    \toprule
    Method & $\text{I}(\gC)$ & Keywords \\
    \toprule
    ConceptSHAP & 0.000 & one, two, gt, new, cl, lt, first, world, mo, last, b, san, tuesday, soccer, time,nhl, Australia, red, bryant\\
    ConceptSHAP & 0.000 & first, new, red, world, Yankees, Australia, giants, nl, as, two, one, ga, last, b, u, tuesday, quo, men\\
    \midrule
    HI-Concept & 0.108 & update, us, fed, wal, op, u, stocks, oil, dollar, delta, hr, ex, Wednesday, world, percent, crude\\
    HI-Concept & 0.151 & red, NBA, football,  Yankees, sports, NFL, team, baseball, olympic, league, game, season, coach\\
    \bottomrule
  \end{tabular}}
\caption{Generated concept keywords  with Average Impact ($\text{I}(\gC)$) from AG-News dataset, BERT model.}
  \label{table:agnews_concepts}
\end{table*}

\subsection{Results on text classification datasets}

The experiment results on text classification datasets are presented in \Cref{table:text-results}. Concepts discovered by the baseline methods lead to tiny changes in prediction outputs, which undermine their reliability. On the contrary, 
on all models, especially pretrained BERT and T5, concepts discovered by HI-Concept induce a larger impact than all the baseline methods, while maintaining faithfulness. This observation consolidates our intuition that pretrained complex language models with more confounding correlations can benefit more from HI-Concept. As a seq2seq text generation model, the pretrained T5 is especially hard to learn for the surrogate models, as the output vocabulary has a class size of 32,128. For better calculation of impact, we simplify outputs by filtering to only the classification classes (e.g.,  words ``Positive'', ``Negative'' for IMDB) and summing all other vocab probabilities as ``Other''. Some models collapse completely in this case. HI-Concept, however, excels in maintaining both faithfulness and a large impact.

To qualitatively examine the discovered concepts, we take an example of BERT on AG-News. In \Cref{table:agnews_concepts}, 2 out of 10 concepts from both HI-Concept and the baseline are shown. For ConceptSHAP, both concepts have low $\text{I}(\gC)$, corresponding to lower impact.
Although vaguely hinting at the category ``Sports'', they consist of words that are less indicative, such as ``one'', ``two'', and ``new''. When looking at HI-Concept discovered topics, the first talks about ``World'', especially in the global finance topic. The second clearly points to the American sports leagues, indicating category ``Sports''. Thus, instead of merely pointing to class information, the concepts discovered contain more information that aligns with human understandable concepts. The $\text{I}(\gC)$ score shown here is also consistent with what humans perceive as important and causal words to the classification, thus indicating the metric's validity. Moreover, in \Cref{app:layer_comparison}, we show more concepts discovered in former layers of BERT, which show that the concepts are not only separable in semantic meanings, but also syntactical information (such as nouns and adjectives).

\begin{figure}
  \centering
    \scalebox{0.70}{  \begin{tabular}{l   p{7.8cm}}
    \toprule
    Method & Visualization \\
    \toprule 
    \multirow{4}{*}{ConceptSHAP} & \hlgreenc{dream team} \hlgreenb{leads} \hlgreenc{spain 44} \hlgreenb{-} \hlgreenc{42 at halftime athens, greece} \hlgreena{- as} \hlgreenb{expected}, \hlgreenc{the u.}\hlgreenb{s. men's} \hlgreenc{basketball team} \hlgreenb{had} its \hlgreenc{hands} full \hlgreena{in a} \hlgreenc{quarterfinal game against spain} \hlgreena{on} \hlgreenb{thursday...} \\
    \midrule
    \multirow{4}{*}{HI-Concept} & \hlgreenb{dream} \hlgreena{team} \hlgreenb{leads} \hlgreenc{spain 44} - \hlgreenc{42} at \hlgreenc{halftime} \hlgreenb{athens}, \hlgreenb{greece} - as expected, the \hlgreenc{u}.s. men's \hlgreenc{basketball} \hlgreenb{team} had its \hlgreenb{hands} \hlgreenc{full} in a \hlgreenc{quarterfinal} \hlgreenb{game} \hlgreenc{against} \hlgreenb{spain} on thursday ...\\
    \bottomrule
  \end{tabular}} 
  \caption{Qualitative comparison from AG-News: ``World'' news misclassified as ``Sports'' by BERT.}
  \label{table:visualization}
\end{figure}

The \textbf{usability} of our method is visualized in  \Cref{table:visualization}, which shows the same failure case ({labeled as ``World'' news but misclassified as ``Sports''}) highlighted with the top concept discovered. ConceptSHAP discovers a top concept related to the keywords ``leads'', ``as expected'', or ``on thursday'', which are not informative as to why the model classified this input as ``Sports''. On the contrary, HI-Concept could precisely point out why: BERT is looking at keywords such as ``dream team'', ``game'', and country names. Such examples show the potential of HI-Concept being used in understanding the model's failure processes, which we further investigate in \Cref{sec:human_study} with a carefully designed human study.

\subsection{Ablation study}

To ensure that the designated 4 objectives behave as expected, we conduct ablation studies for BERT on AG-News and report the results in \Cref{table:ablation}. As observed, eliminating prediction loss leads to a large decrease in RAcc, resulting in an unfaithful model. Thus, even though the model leads to large accuracy changes, the results cannot be trusted. Without auto-encoding loss or regularizer loss, the model has lower performances both in faithfulness and impact. Without causality loss, RAcc is the highest, indicating an accurate reconstruction of the original predictions. However, the discovered set of concepts results in lower output changes. Finally, the full HI-Concept method discovers a set of concepts that both generate high impact and maintain a good level of faithfulness.

\subsection{Human study}
\label{sec:human_study}

To validate that HI-Concept can identify words that are more impactful than the baseline ConceptSHAP, we design the following human study setup: 100 randomly selected examples from AG's testset are shown, where each example consists of the text input and the model's prediction. The annotator is asked to select up to three most causal words for the predicted label. We collect annotations from 4 different annotators proficient in English in order to obtain a diverse set of causal keywords. We consider the keywords selected by the annotators to be the ground-truth and calculate the average impact score ($\text{I}(x_t)$) (\Cref{subsec:metric}) for all unique words (superset) selected by the annotators, with the baseline and our method trained on the penultimate layer of BERT. Details about how the human study is conducted can be found in \Cref{app:human_study}. 

\begin{table}[t]
\centering
\scalebox{0.7}{  \begin{tabular}{lccc}
    \toprule
    Method & $\text{RAcc}$ $\uparrow$  & $\text{I}(\gC)$ $\uparrow$ & $\Delta \text{Acc} \uparrow$ \\
    \toprule
    No Auto-Encoding Loss & 93.46\% & 0.028 & 6.11\% \\
    No Prediction Loss & 68.00\% & 0.035 & \textbf{17.41\%} \\
    No Regularizer Loss & 95.76\% & 0.041 & 6.23\% \\
    No Causality Loss & \textbf{99.92\%} & 0.029 & 2.95\% \\
     \midrule
    HI-Concept & 99.90\% & \textbf{0.058} & 10.54\% \\
    \bottomrule
  \end{tabular}}
\caption{ Ablation on BERT for IMDB with faithfulness (RAcc) and impact ($\text{I}(\gC)$, $\Delta \text{Acc}$) evaluation.}
  \label{table:ablation}
\end{table}

\begin{table}[t]
  \centering
\scalebox{0.7}{  \begin{tabular}{lcccc}
    \toprule
     & \multirow{2}{*}{\# Examples}  & \multirow{2}{*}{Cohen $\kappa$}   & ConceptSHAP & Ours\\
     & & & (Impact) &  (Impact)  \\  
    \toprule
    Total & 100 & 0.41 & 0.41 & \textbf{0.83} \\
    \bottomrule
  \end{tabular}}
  \caption{Human study for causality/usability evaluation.}
  \label{table:human-study}
\end{table}

As shown in \Cref{table:human-study}, the average impact score produced by HI-Concept
is 2 times higher than the baseline, demonstrating that our model selects the right (causal) tokens.  The annotators have a Cohen's Kappa agreement of 0.41, which is considered as moderate agreement \citep{landis1977measurement}. Therefore, even though the annotators prefer slightly different keywords as causal, our model still assigns a higher score to them compared to the baseline, covering causal keywords preferred by all annotators. For example, \Cref{fig:human_vis} shows a correctly classified ``Business'' news article. The human annotators, with a small disagreement, provide 4 unique keywords, which are all highlighted by HI-Concept. 
Such ability to identify causal tokens gives HI-Concept the potential to be used for debugging applications.

\begin{figure}[t]
\centering
    \scalebox{0.65}{  \begin{tabular}{l p{8.4cm}}
     \toprule
     \textbf{HI-Concept} &
    \hlgreenb{google} \hlgreena{shares}, once \hlgreena{devalued}, just may be winners after all \hlgreenc{wall street}, which forced google, the \hlgreenc{internet search engine}, to sharply \hlgreenc{lower} the \hlgreenb{price} of its \hlgreena{shares} in its initial public offering in \hlgreenc{august}, has decided that the \hlgreenc{company} is worth a lot more today than it was then. \\
    \toprule
    \textbf{Human 1} & devalued, shares, price\\
    \textbf{Human 2} & 
    devalued, shares, google \\
    \bottomrule
  \end{tabular}}
\caption{Example sentence in the human study: ``Business'' news correctly classified.}
\label{fig:human_vis}
\end{figure}

\subsection{Hyper-parameter comparisons}

We perform further studies on two hyperparameters: the layer(s) to interpret and number of concepts. We conduct text experiments and evaluate in terms of both impact and concept quality. In this section, we summarize our main findings and refer the readers to \Cref{app:hp_comparison} for experiment details and results (charts and wordcloud visuals).  


For \textbf{layer-wise comparisons}, we experiment on the 3rd, 6th, 9th, and 12th layer respectively, all with 10 concepts. In terms of impact, the intermediate layers (3, 6 and 9) have a higher impact $\text{I}(\gC)$ (around 0.150), while the penultimate layer (12) has a lower $\text{I}(\gC)$ (around 0.058). This is because the concepts discovered at the penultimate layer are sentence-level (using the [CLS] token), while the intermediate layer concepts are token-level. Thus, the sentence-level concepts have less fine-grained control. In terms of topic quality, the later layers tend to discover more coherent concepts, where each concept mostly corresponds to one class label. The beginning layers, on the contrary, tend to discover concepts that are more abstract with mixed class labels. The earlier layers can also discover lexical concepts, such as concepts with only nouns or adjectives. Similarly, \citet{dalvi2021discovering} observe that BERT finds more lexical information in the earlier layers. This interesting observation could lead to future studies in investigating how information flows through different layers in BERT.

For \textbf{number of concepts}, we experiment with 3, 5, 10, 50, and 100 concepts on the penultimate layer. We find that {a concept number close to the number of output classes usually gives higher prediction changes, while increasing the number results in higher recovering accuracy.} When the number of concepts becomes larger, concepts usually become more coherent, although the performance will decrease, as too large a number of concepts introduces more noise into the training process.

\section{Conclusions}

We have proposed a complete framework to derive {impactful} concepts that explain a black-box language model's decisions. Our framework addresses 3 important challenges in NLP explainability: \Ni \textbf{Impactful:} it derives concepts that {generate high output changes and} minimize confounding explanations through a causal loss objective. \Nii \textbf{{Counterfactuals}:} it proposes an innovative substitute to the traditional input counterfactual. By producing latent counterfactuals that are designed to remove features within input texts, we avoid the input space search. The concern of interpreting the hidden activations is also addressed by incorporating visualization methods. 
\Niii \textbf{User-friendly:} as demonstrated with visualizations and human studies, HI-Concept leads to human-friendly explanations in NLP tasks that contain high-level text attributes and semantically meaningful concepts. 




\section*{Limitations}

Regarding potential concerns, HI-Concept only encourages high impact in post-hoc model explanations and should serve as an assistive tool instead of being accepted as ground-truth. 

As a future venue to our work, we believe that our framework will set a good foundation for future research on causal NLP explainability methods, especially those that hope to derive human-friendly explanations. To improve it further, a similar causal objective could be used to address spurious correlations during training. It also has the potential of being carried over to other domains, such as vision or tabular tasks. The high-level attributes in the hidden space can also be used in downstream applications to provide better controllability for the users.

\section*{Ethics Statement}

HI-Concept demonstrates the potential to play an important role in practical scenarios such as debugging and transparency. As AI ethics have become a major concern in real-life applications, such explanations can help users better identify bias and promote fairness.

\bibliography{anthology,custom}
\bibliographystyle{acl_natbib}

\appendix
\clearpage
\begin{center}\large\bfseries
Appendix for ``Explaining Language Models’ Predictions with High-Impact Concepts''
\end{center}
\appendix

\section{Overall method visualization}
\label{app:overall_graph}

\begin{figure}[h]
    \centering
    \includegraphics[width=0.8\linewidth]{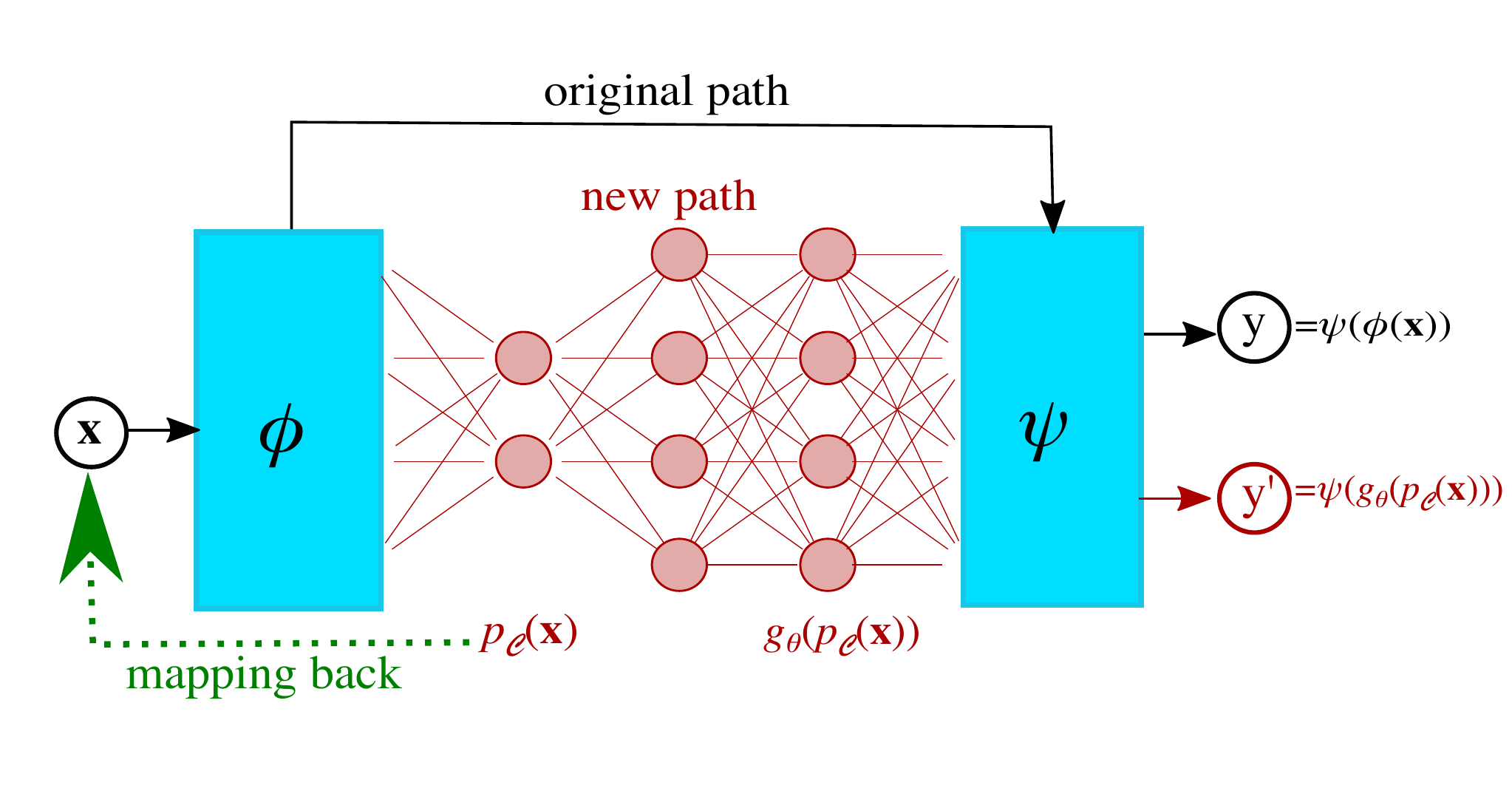}
    \caption{The overall concept generation process.}
    \label{fig:overall}
\end{figure}

\Cref{fig:overall} shows the overall visualization of the concept generation process. The neural network $f$ is divided at the intermediate layer into two parts: $\phi$ and $\psi$. The \textbf{black-arrow} path shows the original neural network prediction process: $y = f(\rvx) = \psi(\phi(\rvx))$, where $\rvx$ is the input and $y$ is the classification output.

To generate concepts at the intermediate layer, instead of feeding $\phi(\rvx)$ directly into $\psi$, we first pass it through a concept network: Firstly, $\phi(\rvx)$ is condensed into concept probabilities $p_{\gC}(\rvx)$ by multiplying the normalized activations $\phi(\rvx)$ with normalized concept vectors $\gC = \{\rvc_1, \dots, \rvc_n\}$ and going through the threshold (TH) function. Then, a 2-layer perceptron $g_\theta$ is used to reconstruct the original activation: $\phi(\rvx) \approx g_\theta(p_{\gC}(\rvx))$. The reconstruction is then passed into $\psi$ to get the prediction $y' = \psi(g_\theta(p_{\gC}(\rvx)))$. To train the network, we use reconstruction loss, regularizer loss, and causality loss.

The \textbf{green path} indicates the mapping back process from concept probabilities $p_{\gC}(\rvx)$ to input tokens in $\rvx = [x_1, \dots, x_t]$. We use the transformer visualization approach \citep{chefer2021transformer} and Grad-CAM \citep{selvaraju2017grad}, which rely on the gradients generated from the red path.

\section{Training details}
\label{app:training}

In practice, we only turn on the causal loss after a certain number of epochs (usually half of the overall number of epochs) to make sure that the {surrogate} model first learns to faithfully reconstruct from the set of concepts before optimizing for the impactful ones. This is because learning the two conflicting objectives at once will usually result in low accuracy. We also note that some contextual information is still needed to maximize the accurate reconstruction of hidden activations $\phi(\rvx)$. Thus, the causality loss is enforced on all concepts except the last one $\rvc_n$, which is used as a `context concept'. During model inference, the last (non-impactful) concept is unused.

After training, we post-process discovered concepts to filter out unused ones. While the number of concepts $n$ is user-selected, as in many topic models, it is an inherent flaw as it requires a certain level of domain expertise. For example, in a movie review dataset with only 2 output classes, if an unfamiliar user sets $n$ to 200, the model will naturally discover many noisy concepts and only a few useful ones. To ensure that the noisy concepts are eliminated, we post-process the concepts and filter out the unused ones (with an impact $I_\text{ind}(\rvc_i)$ close to $0$). Thus, a more desirable number of concepts is returned even if the user provides an overestimate of $n$. In our experiments, we see that, after filtering, the model always achieves a better or same prediction-reconstruction performance as before. However, even with this post-processing, specifying too large a number of concepts can still be dangerous as it harms the concept model's training process.



\section{Toy example}
\label{app:toy}

We conduct experiments on a synthetic (toy) image dataset with ground truth concepts in order to test the validity of our method and confirm the claim that higher confounding effects within the dataset lead to more correlational explanations, thus calling for a more causal explainability approach. Specifically, We extend the toy dataset design of \citet{yeh2020completeness} to make it more realistic by inserting spurious correlations. 

\subsection{Data generation}

As a synthetic setup, at most 15 shapes are randomly scattered on a blank canvas at random locations with random color selections (as noise). For each image sample $x_j$, $\bm{z^j_{\{1:15\}}}$ are binary variables of whether or not a shape is present in $x_j$ with each $\bm{z}^j_s$ sampling from a Bernoulli distribution with probability $0.5$. Then, a 15-class target $\bm{y_j}$ is constructed with respect to whether the first 5 shapes ($\bm{z^j_{\{1:5\}}}$) are present or not with human-designed rules. For example, $\bm{y_1} = \sim (\bm{z_1} \cdot \bm{z_3}) + \bm{z_4}$. A total of $60,000$ examples are generated as the toy dataset using a seed of 0.

The setup mentioned above is, in fact, far away from realistic scenarios, as it does not consider possible confounding. Thus, to make it more realistic, we insert spurious correlations between the pairs $(\bm{z^j_{\{1:5\}}}, \bm{z^j_{\{6:10\}}}), (\bm{z^j_{\{6:10\}}}, \bm{z^j_{\{11:15\}}})$ with a correlation factor $p_\text{cor}$. For example, when $z_1 = 1$, $z_6 = \text{Bernoulli}(p_\text{cor})$; when $z_1 = 0$, $z_6 = \text{Bernoulli}(1-p_\text{cor})$.

\subsection{CNN classification model used for the toy example}

\begin{figure}
    \centering
    \includegraphics[width=0.8\linewidth]{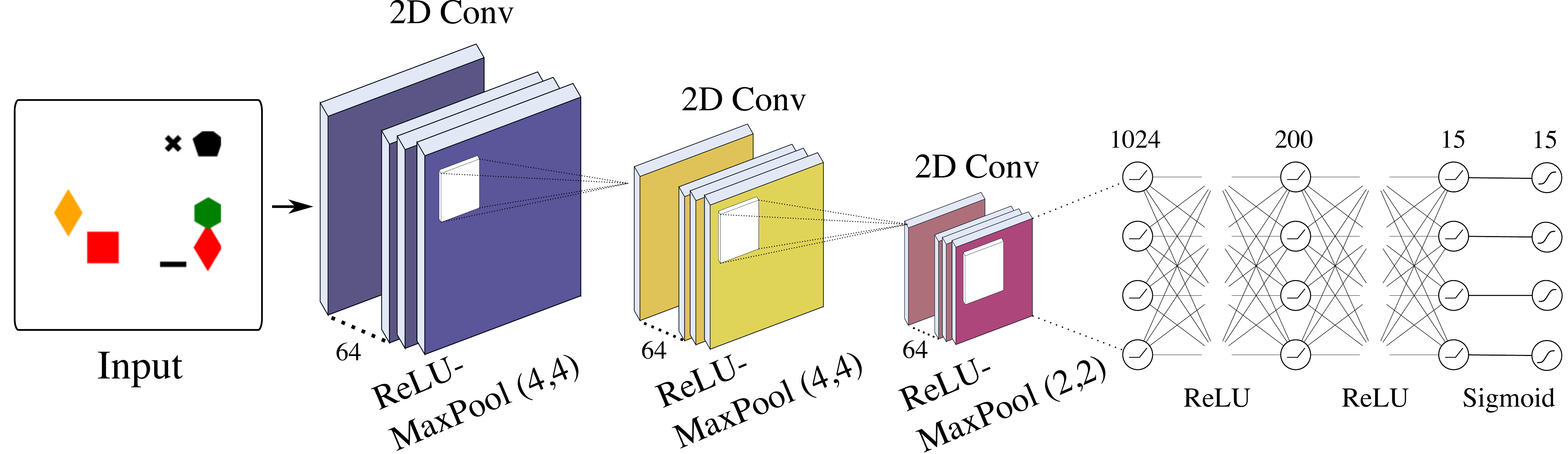}
    \caption{Convolutional Neural Network used for classifying the toy dataset.}
    \label{fig:conv_toy}
\end{figure}

The CNN classification model used for the toy dataset is shown in \Cref{fig:conv_toy}. Specifically, 3 convolutional layers with a kernel size of 5 and 64 output channels were used, each followed by a ReLU activation and max pooling layer. Then, the result is flattened into a linear vector, followed by 2 linear layers and a sigmoid activation function. The output is a 15-dimensional binary classification probability. The model is trained for 100 epochs with an Adam optimizer with learning rate $3e-4$. For reproducibility purposes, the model is initialized and trained with a seed of 0.

\subsection{Visualizations}

\begin{figure}[h]
\minipage{0.49\textwidth}
  \includegraphics[width=\linewidth]{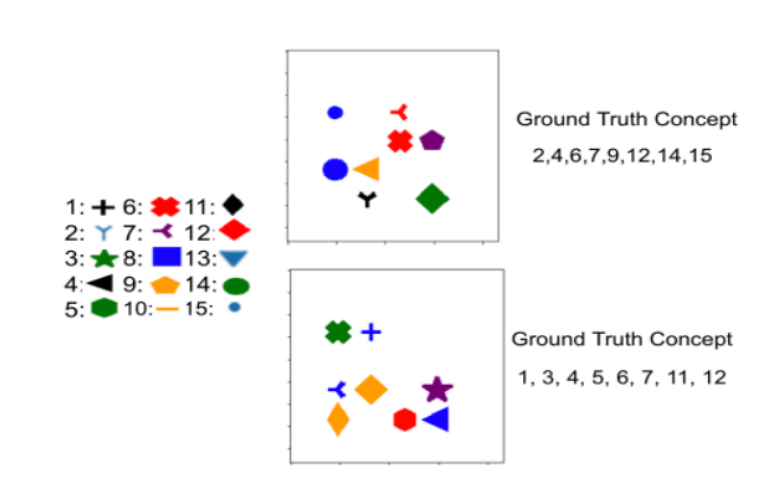}
\endminipage\hfill
\minipage{0.49\textwidth}
  \includegraphics[width=\linewidth]{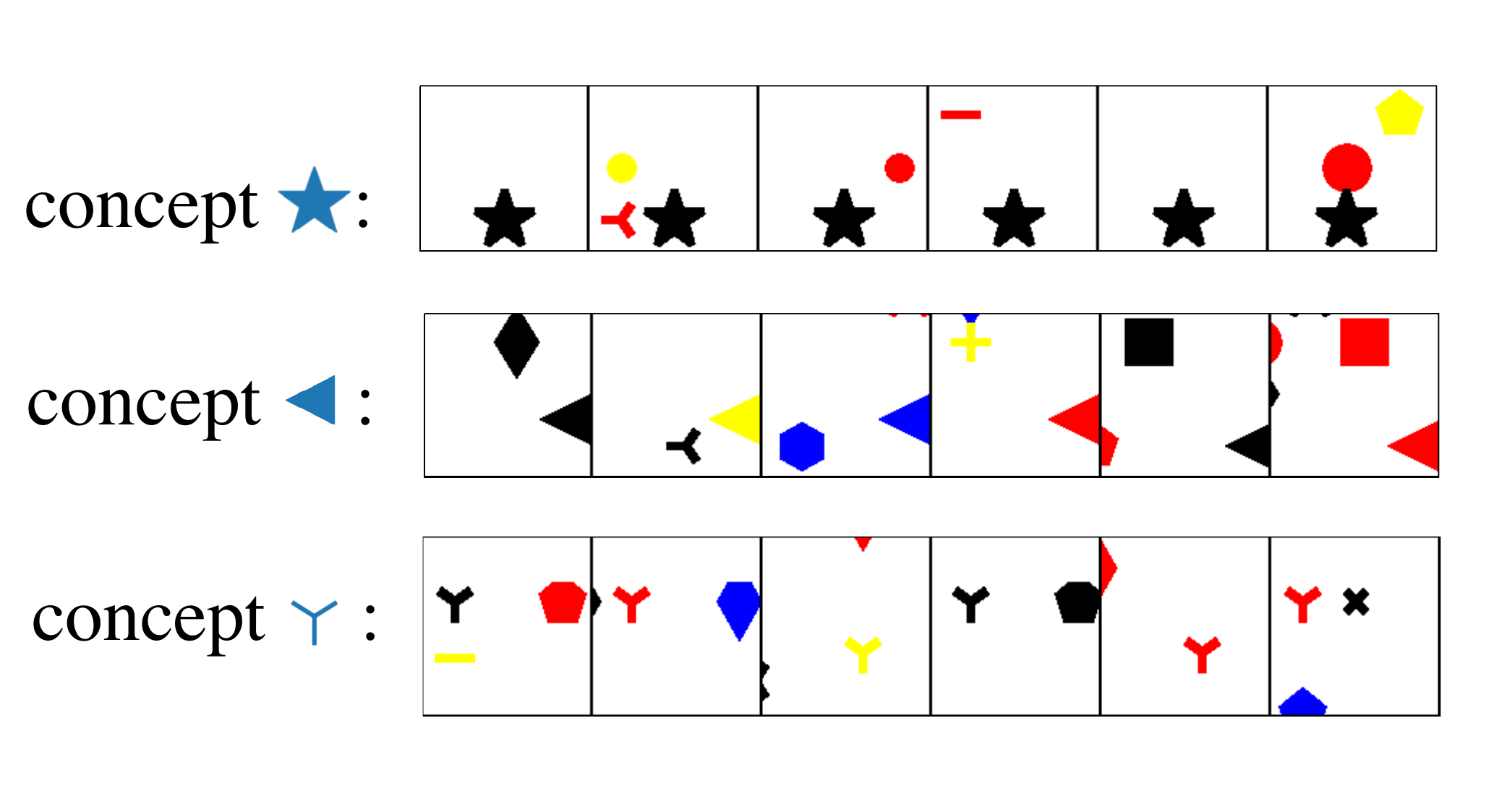}
\endminipage\hfill
\caption{Examples from the toy dataset and concepts discovered.}
\label{fig:toy_dataset}
\end{figure}

As an example visualization, in \Cref{fig:toy_dataset}, two random images from the toy dataset are displayed on the left, while three example concepts discovered by HI-Concept are plotted on the right. We could observe that HI-Concept is able to derive meaningful clusters as concepts, which provide a sanity check for usability of the latent concepts.

\subsection{Results on toy dataset} 

\begin{table}[h]
\centering
\scalebox{0.64}{  \begin{tabular}{lccccc}
    \toprule
    $p_\text{cor}$ & Cls.Acc  & Method & $\text{RAcc}$ $\uparrow$  & $\text{I}(\gC)$ $\uparrow$ & $\Delta \text{Acc} \uparrow$ \\
    \toprule
    \multirow{2}{*}{0.50} & \multirow{2}{*}{95.4\%} & ConceptSHAP & 97.6\% & 0.070 & 6.1\% \\
    & & HI-Concept & \textbf{98.4\%} & \textbf{0.102} & \textbf{9.4\%} (+3.3\%) \\
    \midrule
    \multirow{2}{*}{0.65} & \multirow{2}{*}{99.0\%} & ConceptSHAP & \textbf{99.7\%} & 0.038 & 3.5\% \\
    & & HI-Concept & 99.3\% & \textbf{0.084} & \textbf{6.8\%} (+3.4\%)) \\
     \midrule
    \multirow{2}{*}{0.75} & \multirow{2}{*}{96.1\%} & ConceptSHAP & 98.3\% & 0.069 & 6.0\% \\
    & & HI-Concept & \textbf{98.9\%} & \textbf{0.123} & \textbf{12.16\%} (+6.16\%) \\
    \bottomrule
  \end{tabular}}
\caption{ Faithfulness (RAcc) and Causality ($\text{I}(\gC)$, $\Delta \text{Acc}$) evaluation on the toy dataset. Cls.Acc denotes model's classification accuracy.}
  \label{table:toy-results}
\end{table}

From the results shown in \Cref{table:toy-results}, {we could observe that,} as we increase $p_\text{cor}$ {to mimic an increase in confounding levels in real life}, our HI-Concept consistently outperforms the baseline {by a bigger margin}. {HI-Concept} achieves higher impacts ($\text{I}(\gC)$) and higher accuracy change ($\Delta \text{Acc}$), {while maintaining the best RAcc, indicating faithfulness to the original predictions}. Moreover, we note that the improvement is even stronger in real data experiments, as the added artificial confounding is more complicated in real-life scenarios.

\section{Hyperparameters used}
\label{app:hyperparameters}

\begin{table}
\centering
\scalebox{0.8}{
\begin{tabular}{lllcc}
\toprule
Dataset & Train & {Test} & Label dim. & Avg. size \\
\midrule
Toy (image) & 48k & 12k & 15 & (240, 240) \\
IMDB (text)  & 37.5k & 2.5k & 2 & 215 \\
AG   (text)  & 120k  & 7.6k &  4 & 43  \\
\bottomrule 
\end{tabular}
}
\caption{ A summary of the datasets.}
\label{table:datasets}
\end{table}

For all concept experiments, the following parameters are universally applied as a selected default, which demonstrated better performances during experiments: For regularizer losses, $\lambda_1=0.1$ and $\lambda_2=0.5$. In $\text{TH}(\cdot, \beta)$ function, threshold is set to be $\beta = 0.1 = \frac{1}{n}$, where n is the number of concepts selected. For the top-$N$ neighborhood, $N = \frac{1}{4} \text{BS}$, where $\text{BS}$ is the effective batch size, which we have set as 128 during the experiments. For the masking strategy, we always recommend masking random concepts with a probability of $0.2$ as the optimal strategy, as masking maximum concepts may lead to a highly uneven distribution of $\text{I}(\gC)$ among discovered concepts. 

As all dataset class sizes are small (2 in IMDB/toy or 4 in AG-News), the number of concepts is chosen to be 10 for all experiments. When the number of classes is larger, we recommend choosing a larger number of concepts to ensure a faithful reconstruction of the original input.

For training the concept model, we always use an Adam optimizer with a learning rate of $3e-4$. All models are all trained using 100 epochs. In the HI-Concept models, causal loss is always turned on at half of the overall number of epochs. After turning on causal loss, all parameters are set to untrainable except for the concept vectors, which ensures that the reconstruction ability is not forgotten.

The same hyperparameters are set for the conceptSHAP models, which are also found to generate the optimal performances. The threshold is set to be $\beta = 0.3$, as recommended by the original paper on NLP datasets.

For the causal loss regularizer, $\lambda_{\text{c}}=1$ is set for all experiments, except for $\lambda_{\text{c}}=3$ in the case of IMDB with BERT. A higher $\lambda_{\text{c}}$ will usually lead to a higher output change ($\text{I}(\gC)$ and $\Delta\text{Acc}$), accompanied by a decrease in faithfulness (RAcc). 

To reproduce, all experiments were run with a random seed of 0.

A summary of the datasets is provided in \ref{table:datasets}. IMDB and AG-news are both licensed for non-commercial use.

\section{Run-time}
\label{app:runtime}

\begin{table}
\centering
\label{table:runtime}
\scalebox{0.65}{
\begin{tabular}{lccccc}
\toprule
Dataset & $\beta$-TCVAE & kmeans & PCA & conceptSHAP & HI-Concept \\
\midrule
IMDB  & {475.9} & 37.7 & 0.8 & 199.3 & 227.2\\
AG    & 1525.6  & 15.51 & 2.5 & 1749.65 & {2242.1}  \\
\bottomrule 
\end{tabular}
}
\caption{ A summary of runtime (in seconds) on datasets for BERT.}
\end{table}

As our model optimizes for causality loss, the run-time is slightly longer than the baseline method ConceptSHAP \citep{yeh2020completeness}, but is still short. A summary of runtime is shown in \Cref{table:runtime}. All models shown are run on the GTX 1080Ti graphic card with 12 GB memory. Generally, as post-hoc explainability methods, the runtimes are very light and, therefore, a concern that is less important than the model quality. For example, on a dataset of size 50k such as IMDB, it only takes 227.2 seconds (3.8) minutes to train our HI-Concept model.

\section{Human study setup}
\label{app:human_study}
\begin{figure}
    \centering
    \includegraphics[width=0.95\linewidth]{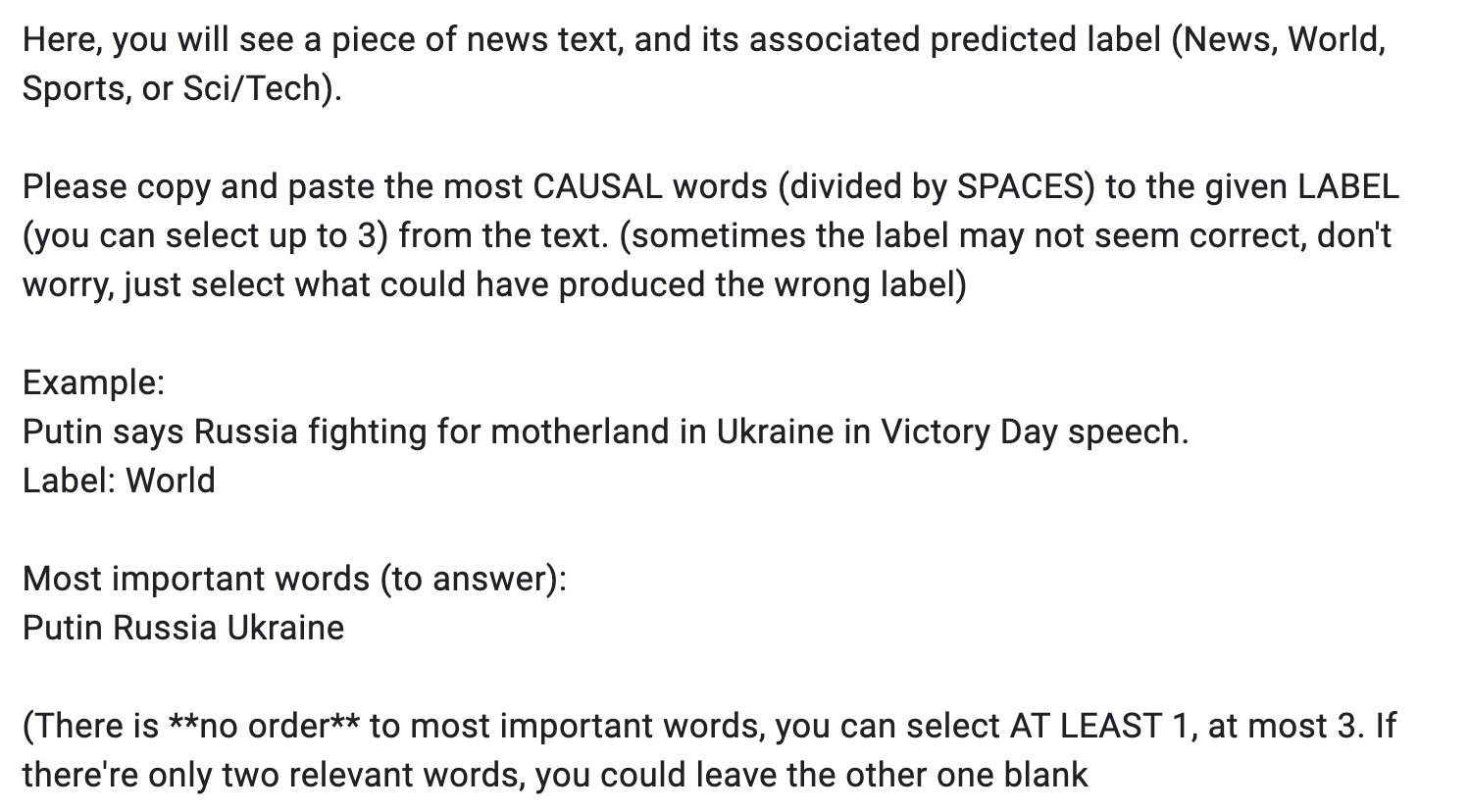}
    \caption{Human study instructions with a demonstration.}
    \label{fig:human_study_instructions}
\end{figure}

\begin{figure}
    \centering
    \includegraphics[width=0.95\linewidth]{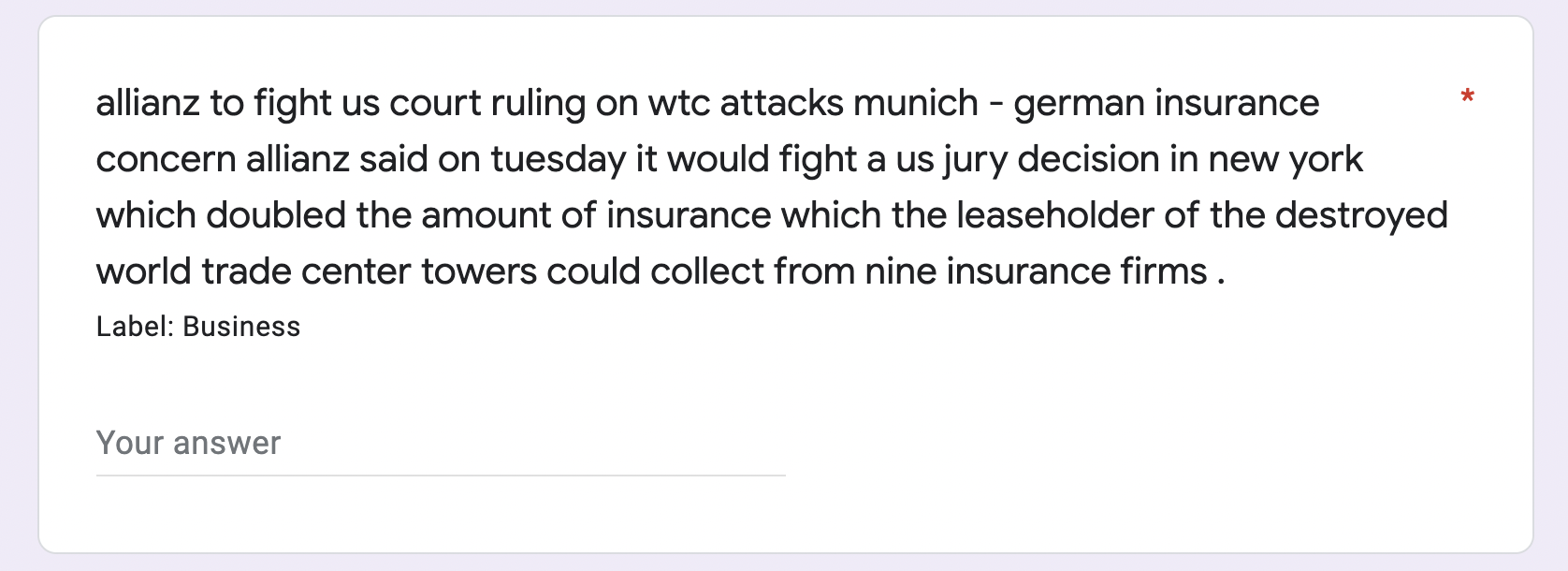}
    \caption{Human study question and answer.}
    \label{fig:human_study_qa}
\end{figure}

For the human study, 100 examples are randomly selected from the test set $\gD_{\text{test}}$. The questionnaire takes the format of a google form, where the instructions in \Cref{fig:human_study_instructions} are shown to the participants. An example question looks like the one in \Cref{fig:human_study_qa}. For the 100 questions repeated twice, 4 volunteers (Ph.D. students) have answered them. The volunteers are all proficient in English. The volunteers report an average time of 30 minutes for answering 50 questions. As the volunteers are working also in AI-related areas and are briefed about the purpose and usage of survey data beforehand, they understand fully the data collection and usage. Thus, implicit consent is granted by participation.

\section{Hyperparameter comparisons}
\label{app:hp_comparison}

The proposed method of HI-Concept includes many tunable hyperparameters, including the top-N neighborhood, threshold, etc. While these parameters are set at the default mentioned in \Cref{app:hyperparameters}, there are two hyperparameters that users can customize the most: the layer to interpret at and number of concepts . To better understand how these two parameters may affect the generated concepts, we conduct comparisons on both. We evaluate in terms of impact and topic quality. For impact, we have reported the number of effective concepts left after post-processing, the recovering accuracy (RAcc), the Average Impact ($\text{I}(\gC)$), and the induced change in accuracy ($\Delta \text{Acc}$). For topic quality, we have reported coherence scores, including averaged Pointwise Mutual Information (PMI) (c\_uci score), normalized PMI (c\_npmi score), c\_v score which measures how often the topic words appear together in the corpus, and word2vec similarity \citep{roder2015exploring}. 

The following comparisons are all conducted on the AG-news dataset with BERT, where the other hyperparameters mentioned in \Cref{app:hyperparameters} stay the same. 

\subsection{Layer-wise comparison}
\label{app:layer_comparison}

\begin{figure*}[h]
\includegraphics[width=\linewidth]{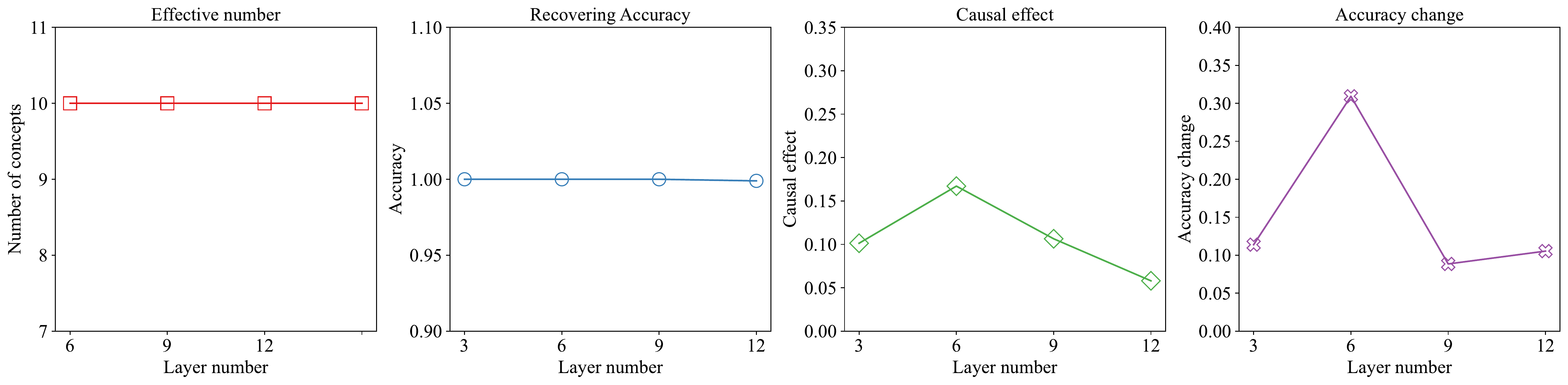}
\caption{Layer-wise effective number of concepts, RAcc $\uparrow$, $\text{I}(\gC)$ $\uparrow$, and $\Delta$ Acc $\uparrow$.}
\label{fig:layerwise_causal}
\end{figure*}

To compare what each layer discovered, as BERT has 12 layers, we experimented on the 3rd, 6th, 9th, and penultimate layer respectively, all with 10 concepts. 

Quantitatively, we plotted out the effective number of concepts, recovering accuracy, impact and accuracy change in \Cref{fig:layerwise_causal}. All layers demonstrate similar performances in recovering accuracy, which is close to 100\%. The intermediate layers, especially the 6th layer, produce a higher average impact and recovering accuracy. This is because the intermediate layers discover concepts on the token-level, while the penultimate layer concepts are sentence-level (on the [CLS] token). Thus, the token-level concepts will have more fine-grained control. 
\begin{figure*}[h]
\minipage{0.32\textwidth}
  \includegraphics[width=\linewidth]{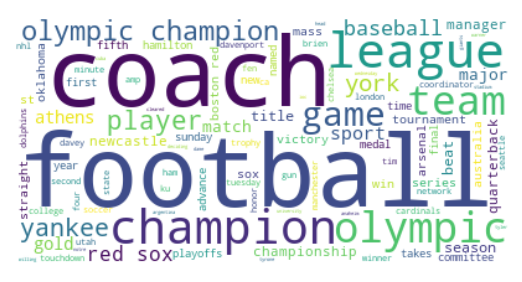}
\endminipage\hfill
\minipage{0.32\textwidth}
  \includegraphics[width=\linewidth]{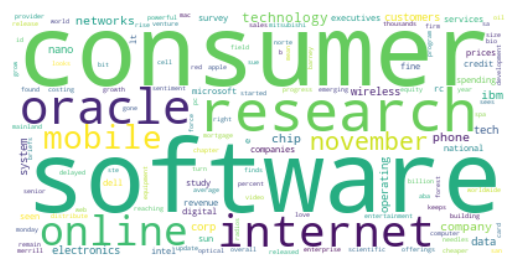}
\endminipage\hfill
\minipage{0.32\textwidth}%
  \includegraphics[width=\linewidth]{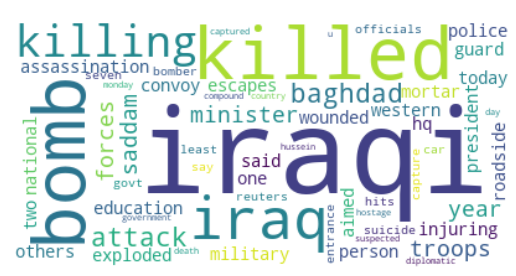}
\endminipage
\caption{Wordclouds of concepts generated on the 12th layer, including a sports concept, a technology concept, and a political concept.}
\label{fig:layer_12_wordclouds}
\end{figure*}

\begin{figure*}[h]
\minipage{0.32\textwidth}
  \includegraphics[width=\linewidth]{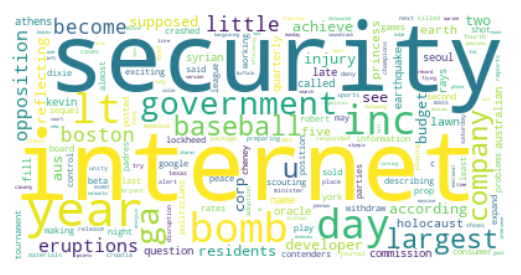}
\endminipage\hfill
\minipage{0.32\textwidth}
  \includegraphics[width=\linewidth]{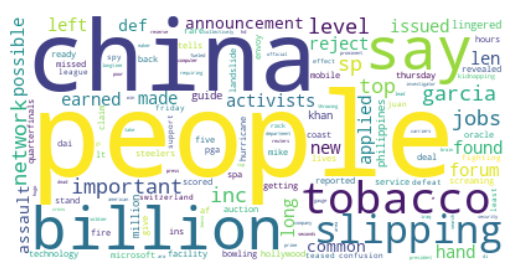}
\endminipage\hfill
\minipage{0.32\textwidth}%
  \includegraphics[width=\linewidth]{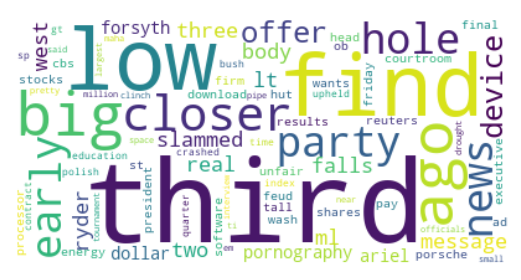}
\endminipage
\caption{Wordclouds of concepts generated on the 9th layer, including a government concept, a China concept, and an Adjective (mostly) concept.}
\label{fig:layer_9_wordclouds}
\end{figure*}

Qualitatively, we plotted some wordclouds of the keywords in discovered concepts in \Cref{fig:layer_12_wordclouds} and \Cref{fig:layer_9_wordclouds}. From \Cref{fig:layer_12_wordclouds}, we could see that, in the penultimate layer, concepts are more concentrated on each class. For example, the first concept would correspond to the class ``Sports'', the second to ``Sci/Tech'', and the third to ``World'' news. The emphasis on events is also clearer, such as the third one talking about the Iraq War. However, When we move to earlier layers, the concepts’ class labels are more mixed together. In \Cref{fig:layer_9_wordclouds}, the first concept concerns government, which includes terms such as ``government'', ``internet'', ``security'', ``bomb'', ``baseball'', etc. It could, however, correspond to many class labels, such as ``Sci/Tech'', ``World'', or even ``Sports''. Similarity, the second concept talks about China, including ``china'', ``billion'', ``people'', ``activitists'', ``announcement'', etc. The third concept is interesting as it covers mostly adjective words which do not seem to correlate too much in semantic meanings, such as ``low'', ``big'', ``closer'', and ``third''. Similar observations are also confirmed in papers such as \citep{dalvi2021discovering}, which derives concepts using agglomerative hierarchical clustering combined with human annotations in BERT latent representations. They observe that BERT finds more lexical information in the earlier layers.

\begin{figure*}[h]
\includegraphics[width=\linewidth]{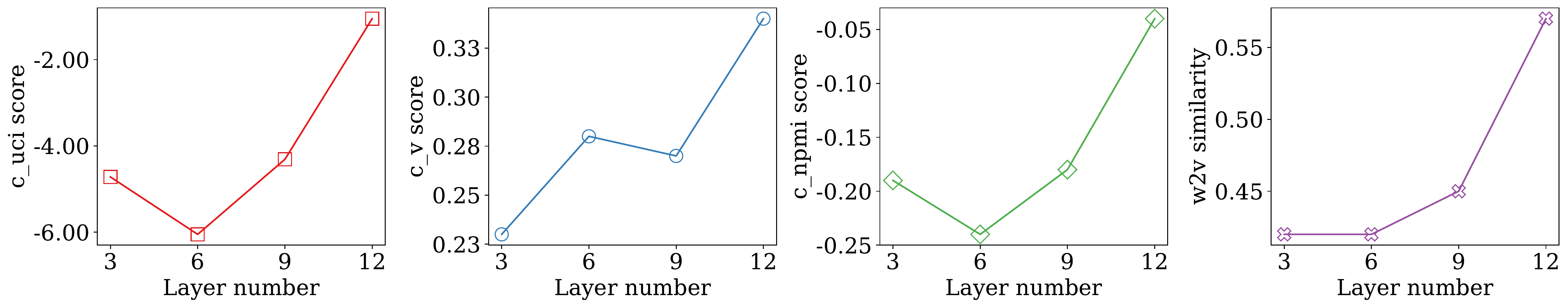}
\caption{Layer-wise Topic Coherence Comparison.}
\label{fig:layerwise_coherence}
\end{figure*}

In terms of topic quality, we evaluated the concept keywords using coherence metrics. As shown in \Cref{fig:layerwise_coherence}, all coherence scores showed a general trend of concepts becoming more coherent as the layer number increases. The conclusion is consistent with the wordcloud visualizations.

Thus, in real-life debugging scenarios, we recommend using the penultimate layer, which will find more coherent topics. However, there could be continued work to discover information learned in the prior layers and to investigate how information flows through layers in a hierarchical way.

\subsection{Number of concepts}

In the penultimate layer of BERT, we experiment with 3, 5, 10, 50, and 100 concepts.

\begin{figure*}[h]
\includegraphics[width=\linewidth]{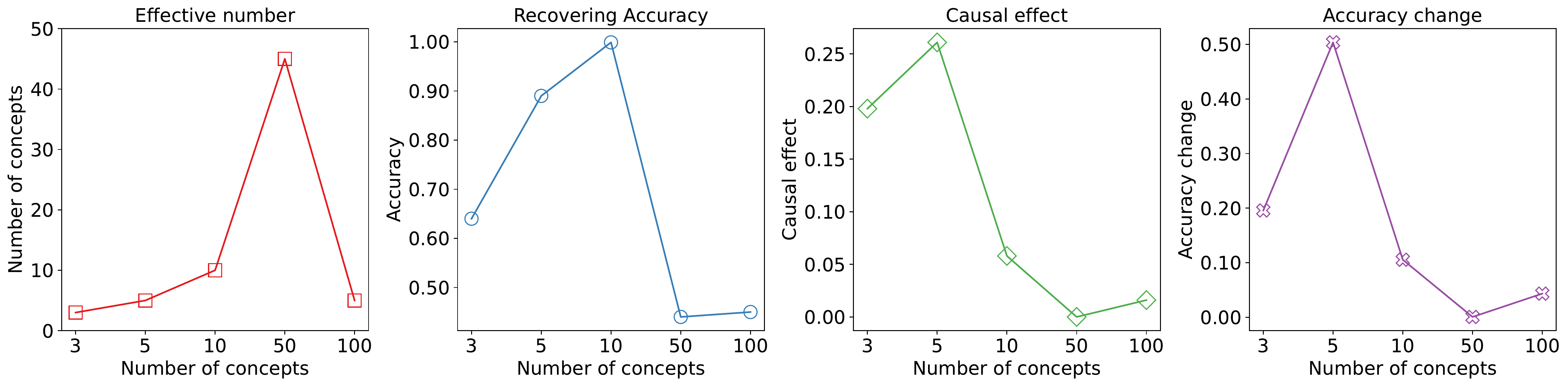}
\caption{Concept-wise effective number of concepts, RAcc $\uparrow$, $\text{I}(\gC)$ $\uparrow$, and $\Delta$ Acc $\uparrow$.}
\label{fig:conceptwise_causal}
\end{figure*}

\begin{figure*}[h]
\includegraphics[width=\linewidth]{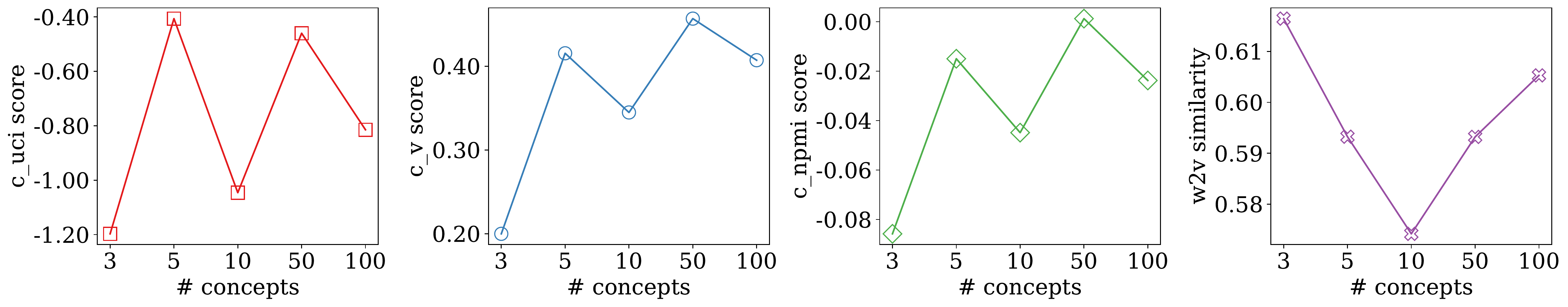}
\caption{Concept-wise Topic Coherence Comparison.}
\label{fig:conceptwise_coherence}
\end{figure*}

From \Cref{fig:conceptwise_causal}, we could see that the performance is very dependent on the number of concepts. The effective number of concepts, recovering accuracy, average impact, and accuracy change all appear to be elbow-shaped. In this case, 5 concepts provided the highest impact on output predictions, as it is close to the number of classes (4) in the AG-News dataset. Increasing the number of concepts to 10 would yield a better recovering accuracy. As the number of concepts increases to 50 and 100, we observe that the model fails to learn completely. In practice, we have often observed the best number to be positively correlated with the number of dataset classes. In other words, a dataset with more classes will require a higher number of concepts for faithful reconstruction. In terms of topic coherence, we could observe from \Cref{fig:conceptwise_coherence} that the topic coherence scores usually oscillate, but mostly display a generally upward trend of becoming more coherent as the number of concepts increases.

\section{Classification models used for text experiments}


\subsection{Transformer classification model trained from scratch}
The self-trained transformer model used during text experiments follows a simple structure: the input text is truncated to max length 512 and passed to an embedding layer of dimension 200. Then, the embeddings are passed through a positional encoding layer with dropout rate 0.2. Then, 6 transformer layers follow with a hidden dimension of 200 and 2 heads. Finally, we mean pool the transformed embeddings and pass through a linear classifier head. The linear outputs are activated with a Sigmoid function to produce class probabilities.

To train the transformer model, we use either the IMDB or AG-News dataset. We train for 10 epochs with a batch size of 128 and an Adam optimizer with learning rate $3e-4$. We also use a learning rate step scheduler with step size 1 and gamma of 0.95.

\begin{table}[h]
  \centering
  \label{table:hyperparameters_bert}
\scalebox{0.8}{  \begin{tabular}{lcc}
    \toprule
     Dataset & AG-News & IMDB\\
    \toprule
     LR & $5e-5$ & $3e-4$ \\
     train BS & 8 & 8 \\
     eval. BS & 16 & 16 \\
     seed & 42 & 42 \\
     optimizer & Adam & Adam \\ 
     & $\text{betas}=(0.9,0.999)$ & $\text{betas}=(0.9,0.999)$\\
     & $\text{epsilon}=1e-8$ & $\text{epsilon}=1e-8$\\
     LR scheduler & linear & linear \\
     warmup steps & 7425 & 1546 \\
     training steps & 74250 & 15468\\
    \bottomrule
  \end{tabular}}
  \caption{ Hyperparameters for finetuning BERT model.}
\end{table}

\subsection{Pretrained and finetuned BERT model}

For AG-News, we take the finetuned version of bert-base-uncased model on huggingface: ``fabriceyhc/bert-base-uncased-ag\_news''. For IMDB, we finetuned by ourselves on the bert-base-uncased model. The hyperparameters used for both finetuning are reported in \Cref{table:hyperparameters_bert}, where LR stands for learning rate and BS stands for batch size. 

The huggingface code and models are all licensed under Apache 2.0, which allows for redistribution and modification. Similarly, the codebase used for replicating the visualization method \citep{chefer2021transformer} and the baseline method \citep{chen2018isolating} are licensed under the MIT license, which allows for redistribution of the code.

\end{document}